\documentclass{article}
\PassOptionsToPackage{numbers, compress}{natbib}

\usepackage[main, final]{neurips_2026}

\usepackage[utf8]{inputenc} 
\usepackage[T1]{fontenc}    
\usepackage{hyperref}       
\usepackage{url}            
\usepackage{booktabs}       
\usepackage{amsfonts}       
\usepackage{nicefrac}       
\usepackage{microtype}      
\usepackage{xcolor}  

\usepackage{amsmath}
\usepackage{amssymb}
\usepackage{mathtools}
\usepackage{amsthm}
\usepackage{algorithm}
\usepackage{algpseudocode}
\usepackage{enumitem}
\usepackage{multirow}

\usepackage[capitalize,noabbrev]{cleveref}
\usepackage{subcaption}

\newcommand{\data}{\mathcal{D}_n}
\newcommand{\datab}{\mathcal{D}_n^b}
\renewcommand{\H}{\mathrm{H}}
\newcommand{\thetab}{\hat{\theta}_b}
\newcommand{\fisher}{\mathcal{I}^{-1}(\theta_0)}
\newcommand{\I}{\mathrm{I}}
\newcommand{\E}{\mathbb{E}}
\newcommand{\cN}{\mathcal{N}}
\newcommand{\var}{\mathbb{V}\textrm{ar}}
\newcommand{\xt}{X_\mathrm{test}}
\newcommand{\yt}{Y_\mathrm{test}}
\newcommand{\phat}{\hat{p}(\xt;\theta)}
\newcommand{\phatb}{\hat{p}(\xt;\thetab)}
\newcommand{\phatk}{\hat{p}_k(\xt;\theta)}
\newcommand{\phatkb}{\hat{p}_k(\xt;\thetab)}
\newcommand{\1}{{\rm 1}\kern-0.24em{\rm I}}
\renewcommand{\P}{\mathbb{P}}
\DeclareMathOperator*{\argmax}{argmax}

\theoremstyle{plain}
\newtheorem{theorem}{Theorem}[section]

\theoremstyle{definition}

\theoremstyle{remark}

\newcommand{\citetay}[1]{\citeauthor{#1} (\citeyear{#1})}

\title{Deep Ensembles for Epistemic Uncertainty: A Frequentist Perspective}

\author{%
  Anchit Jain\\
  EECS, MIT\\
  Cambridge MA, USA \\
  \texttt{ajain625@mit.edu} \\
  \And
  Stephen Bates \\
  EECS, MIT \\
  Cambridge MA, USA \\
  \texttt{s\_bates@mit.edu}\\
}

\begin{document}

\maketitle

\begin{abstract}
  Decomposing prediction uncertainty into aleatoric (irreducible) and epistemic (reducible) components is critical for the reliable deployment of machine learning systems. While the mutual information between the response variable and model parameters is a principled measure for epistemic uncertainty, it requires access to the parameter posterior, which is computationally challenging to approximate. Consequently, practitioners often rely on probabilistic predictions from deep ensembles to quantify uncertainty, which have demonstrated strong empirical performance. However, a theoretical understanding of their success from a frequentist perspective remains limited. We address this gap by first considering a bootstrap-based estimator for epistemic uncertainty, which we prove is asymptotically correct. Next, we connect deep ensembles to the bootstrap estimator by decomposing it into data variability and training stochasticity; specifically, we show that deep ensembles capture the training stochasticity component. Through empirical studies, we show that this stochasticity component constitutes the majority of epistemic uncertainty, thereby explaining the effectiveness of deep ensembles.
\end{abstract}

\section{Introduction}
While machine learning systems are increasingly accurate in tightly-controlled settings, a key challenge to their deployment in more complex, real-world environments is uncertainty awareness---algorithms should be able to accurately report on their level of confidence or reliability in different conditions. Moreover, there are different underlying reasons for uncertainty. Uncertainty may be due to inherent unpredictability in a population with the current feature set (aleatoric or irreducible uncertainty) or instead due to limited training data (epistemic or reducible uncertainty). Distinguishing these types of uncertainty is important in order to take the correct downstream action.  High aleatoric uncertainty indicates that the current prediction task is inherently challenging, and we cannot have high confidence in any one prediction unless we are able to collect richer data (not just more data). On the other hand, high epistemic uncertainty signals that collecting more training data in this region is likely to improve the model, which makes it suitable for guiding model development and active learning \citep{houlsby_bald, gal_deep_al, epig}. We focus on this quantity in this work.

A popular measure of epistemic uncertainty is the mutual information (MI) between model parameters and the target label which quantifies the expected reduction in uncertainty from observing new data \citep{houlsby_bald, uq_decomp_depeweg}. MI is a fundamentally Bayesian quantity that requires access to the posterior distribution over model parameters, which generally renders its exact calculation intractable. As such, substantial effort has gone into developing approximate Bayesian computational techniques~\citep{mackay_bnns, vi_for_dnns, dnn_mcmc, dnn_ep, dropout}, but they are often difficult to scale, compromise predictive accuracy or have weak Bayesian justification. 



In light of these challenges, practitioners often use deep ensembles \citep{deep_ensembles} to obtain uncertainty estimates. Deep ensembles proceed by generating an ensemble of models by only varying the randomization in the training procedure (e.g. initial weights, data shuffling etc.). Previous works have noted that probabilistic predictions from the deep ensemble reliably quantify uncertainty and exhibit strong performance on downstream tasks \citep{deep_ensembles, random_seeds_uq_eval_ovadia, random_seeds_uq_eval_ashukha}. While Bayesian interpretations have been proposed \citep{ensembles_are_bayesian}, a theoretical frequentist understanding behind the success of deep ensembles is lacking. Furthermore, various sources such as limited data, misspecified models or noisy optimization contribute to the uncertainty \citep{aleatoric_epistemic_review} and estimates may only account for some of these. Therefore, to correctly use an uncertainty estimate, it is crucial to understand the sources it captures. Our work aims to provide a frequentist perspective on the epistemic uncertainty captured by deep ensembles.

Our analysis proceeds in three parts. Firstly, we provide an asymptotic expansion for MI which links it to the Fisher information and shows how epistemic uncertainty differs from frequentist variance. We use this to motivate the construction of a bootstrap estimator which targets the MI. Secondly, since the epistemic uncertainty within a deep ensemble arises only due to the randomness in the training procedure, we decompose the bootstrap estimator into components arising due to data sampling and training randomness to facilitate comparison. Finally, through empirical studies, we note that deep ensembles capture the portion of epistemic uncertainty due to stochastic optimization, which in turn appears to be the majority component, lending theoretical justification to their success from a frequentist perspective. 
Our key contributions are---
\begin{itemize}[leftmargin=*]
    \item \textbf{Link between mutual information and frequentist Fisher information:} We provide an asymptotic expansion for MI and show how it depends not only on the variance in model predictions as measured by the Fisher information, but also on the degree of randomness in the true data generating process.
    \item \textbf{Contributions to MI from data variability and training randomness:} We use the asymptotic expansion to construct a bootstrap estimator of MI. To allow comparison with deep ensembles, we show how this bootstrap estimator separates into components arising from training stochasticity and variability in data sampling.
    \item \textbf{Empirical comparison with deep ensembles:} We conduct empirical studies across a variety of deep learning tasks that show that training stochasticity is dominant over variability in data sampling, with deep ensembles capturing the latter.
\end{itemize}
\subsection{Illustration}
\begin{figure*}[t]
    \centering
    \includegraphics[width=\linewidth]{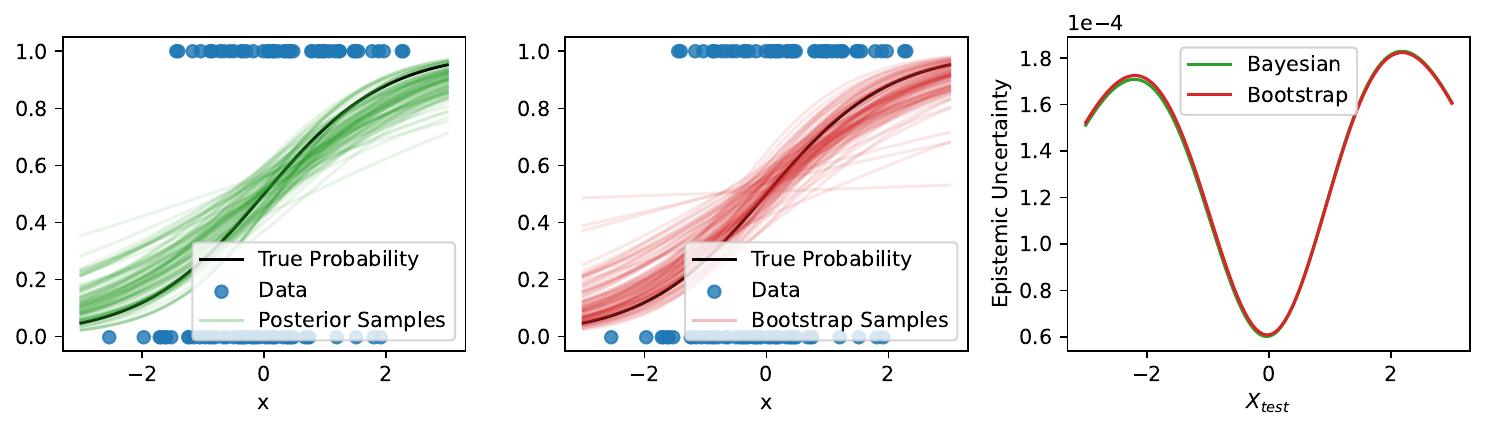}
    \caption{\textbf{Teaser Simulation:} \textit{(left)} Training data, true label probability and posterior sample predictions. \textit{(center)} Predictions from MLEs over bootstrapped datasets. \textit{(right)} Epistemic uncertainty calculated through Bayesian inference (MCMC) versus the bootstrap estimate.} 
    \label{fig:illustrative}
\end{figure*}
To provide the key intuition motivating the bootstrap estimator, we provide the following simulation. Consider a binary logistic regression model, $\P(y=1|x,\theta) = 1/(1+e^{-\theta_1x-\theta_0})$. We sample the features $x$ from a standard normal and also impose standard normal priors on $\theta_0, \theta_1$. Fig.~\ref{fig:illustrative}\textit{(left)} visualizes the data and the true model along with 100 posterior samples from the true Bayesian posterior, obtained using a Markov chain Monte Carlo (MCMC) algorithm. Next, we calculate maximum likelihood estimates (MLEs) for the parameters across 100 bootstrapped datasets and show these in Fig.~\ref{fig:illustrative}\textit{(middle)}\textemdash note the remarkable similarity in the posterior and bootstrap samples. Finally, we use these MCMC and bootstrap samples to form our estimate of the epistemic uncertainty as per Algorithm~\ref{alg:bootstrap} noting the excellent agreement in Fig.~\ref{fig:illustrative}\textit{(right)}. We leverage this connection to study Bayesian MI using bootstrapping, which then allows a comparison with deep ensembles.
\section{Background}
\label{sec:background}
\label{sec:mi_as_eu}

\textbf{Mutual Information as Epistemic Uncertainty:} A popular uncertainty decomposition \citep{mackay, houlsby_bald, uq_decomp_depeweg} leverages information-theoretic quantities to propose measures of aleatoric and epistemic uncertainty. Let the random variables $Y$ and $X$ denote the labels and features respectively in a supervised learning setup. Let $\data$ be the data comprising of $n$ realizations of $(Y,X)$. Further, let $\theta$ be the random variable denoting the parameters of a prediction model. Information theory uses the Shannon entropy $\H(Y)$ to measure uncertainty in $Y$, and defines mutual information as
\begin{equation}
\label{eq:uq_decomp}
    \underbrace{\I(Y;\theta|X,\data)}_{\textrm{Epistemic}} = \underbrace{\H(Y|X,\data)}_{\textrm{Total}} - \underbrace{\H(Y|\theta,X,\data)}_{\textrm{Aleatoric}}.
\end{equation}
The quantification of epistemic uncertainty as the mutual information (MI) between the (test) label and parameters is conceptually pleasing since Eq.~\ref{eq:uq_decomp} reveals it to be the reduction in label uncertainty from knowing the model parameters. This MI is a fundamentally Bayesian object since it requires $\theta$ to be random.

\textbf{Bootstrap:} The original bootstrap procedure \citep{efron_original_bootstrap} was proposed to estimate the variance of an estimator under different draws of the data.
Consider a setting where there are no repeated data points. Bootstrap first samples weights from a symmetric multinomial distribution with all event probabilities $1/n$. These weights are used to reweight the original data to create a ``bootstrapped dataset'' over which the estimator is recomputed. This procedure is replicated $B$ times to obtain $B$ ``bootstrapped estimates". The key intuition behind the bootstrap\textemdash the ``plug-in'' principle\textemdash proposes that the distribution of these bootstrapped estimates approximates the distribution of the original estimator under redraws of the data. As such, we can obtain an estimate of the variance of our estimator as the empirical variance of the bootstrapped estimates.
\\
We can also instead draw weights from a symmetric Dirichlet distribution \citep{rubin_bayesian_bootstrap}. We hypothesize that in a deep learning setting, these weights are more suitable since it allows the neural network to still see all the training instances and hence capture the training data better than multinomial weights which omit training points entirely. For our paper, we thus use Dirichlet weights for our experiments but emphasize that our theorems hold for multinomial weights too (Appendix~\ref{sec:app:proofs}).

\textbf{Deep Ensembles:} Deep ensembles \citep{deep_ensembles} are a simple but powerful uncertainty quantification technique for deep neural networks (DNNs). A deep ensemble is obtained by training multiple models with the exact same architecture and training data but with different random seeds controlling stochasticity in the training procedure (different parameter initializations, different data shuffling when using stochastic gradient descent etc.). The deep ensemble model prediction is simply the average of predictions of member models. They have shown success at a variety of uncertainty quantification tasks \citep{deep_ensembles, random_seeds_uq_eval_ovadia, random_seeds_uq_eval_ashukha}.

\section{Algorithm and Theoretical Results}
\label{sec:algorithm_theoretical_Results}
\subsection{Problem Setup}
We now specialize the previous notation to our exact setting. Consider a $K$-class supervised classification problem with labels $Y\in\{1,...,K\}$. Denote the training data by $\data = \{(X_i,Y_i), i=1,...,n\}$. We wish to evaluate the epistemic uncertainty in the prediction for the label $\yt$ given the feature $\xt$. We denote our probability predictions for $\yt$ for a model parametrized by $\theta\in\Theta$ by $\phat$. Further, denote by $\phatk$ the prediction probability assigned to the $k^{\textrm{th}}$ class. The model is specified such that $0<\phatk<1$ and $\sum_{k=1}^K\phatk=1$. Finally, we also assume that the model is well-specified, i.e., there exists a $\theta_0\in\Theta$ such that $(X_i,Y_i)\stackrel{iid}{\sim}P_X(x)\hat{p}_y(x;\theta_0)=:P_{\theta_0}$, where $P_X$ denotes the marginal distribution of $X$.

In the Bayesian setup, we have a prior $p(\theta)$ over the parameters and we denote by $p(\theta|\data)$ our posterior. We use the notation $\E_Z[.]$ to denote expectations taken with respect to the random variable $Z$. As per Sec.~\ref{sec:mi_as_eu}, the epistemic uncertainty is
\begin{align*}
\I\left(\yt;\theta|\xt,\data\right) & = \H\left(\yt|\xt,\data\right) - \H\left(\yt|\theta,\xt,\data\right) \\
&=\H\left(\E_{\theta|\data}\left[\phat\right]\right) - \E_{\theta|\data}\left[\H(\phat)\right].
\end{align*}

\subsection{Asymptotics of Mutual Information}
We begin by deriving the asymptotic behavior of mutual information, connecting it with the Fisher information and frequentist sampling variation. This connection will allow us to repurpose frequentist methods for estimating variation to instead target the mutual information. 

\begin{theorem}
\label{thm:asymptotics}
Under the assumptions needed for Bernstein von-Mises to hold and uniform integrability of the predicted probabilities (\ref{app:assumption:thm1:start}-\ref{app:assumption:thm1:final} in Appendix~\ref{sec:app:proofs}) we have
the asymptotic expansion
\begin{equation}
\label{eq:mi_expansion}
    \I(\yt;\theta|\xt,\data) = \frac{1}{2n}\sum_{k=1}^K \frac{\sigma_k^2}{\hat{p}_k(\xt;\theta_0)} +o_p(n^{-1})
\end{equation}
$\forall \ \xt$, where $\sigma_k^2$ is
\begin{equation*}
    \sigma_k^2 = \left[\frac{\partial\hat{p}(\xt;\theta_0)}{\partial\theta}^{T}\fisher\frac{\partial\hat{p}(\xt;\theta_0)}{\partial\theta}\right]_{k,k}
\end{equation*}
for $\fisher$ being the inverse Fisher information of the model at $\theta_0$.
\end{theorem}
\textbf{Proof Sketch:} Our analysis proceeds by Taylor expanding the entropy function about the mean posterior predicted probability. We then apply the functional delta method to the Bernstein von-Mises theorem to obtain the limiting law of our predictions. Recall that the Bernstein von-Mises theorem holds under correct model specification, sufficient smoothness, identifiability, and feasibility of $\theta_0$ under the prior (e.g. Theorem 10.1 of \citetay{vdv}). Finally, we use uniform integrability to obtain the limiting posterior moments. We defer the full proof to Appendix~\ref{sec:app:proofs}.

We briefly digress to comment on the theorem and provide better intuition for epistemic uncertainty as measured by mutual information.
Firstly, the theorem shows that mutual information decreases linearly in the number of samples. Next, it connects mutual information to the Fisher information and perhaps surprisingly shows how the first order asymptotics are governed by the inverse Fisher information divided by the true conditional probabilities.

Better intuition can be gained into the first order asymptotic term by considering the binary classification case in which the first order term simplifies to 
\begin{equation*}
    \frac{\left(\partial\hat{p}_1(\xt;\theta_0)/\partial\theta\right)^{T}\left(n^{-1}\fisher\right)\left(\partial\hat{p}_1(\xt;\theta_0)/\partial\theta\right)}{2\hat{p}_1(\xt;\theta_0)(1-\hat{p}_1(\xt;\theta_0))}
\end{equation*}
as noted in~\citetay{mackay}.
Recall that $n^{-1}\fisher$ is the (asymptotic) variance of the MLE. The numerator in the expression above can then be interpreted as the variance of the model prediction at $\xt$. $1/2p(1-p)$ achieves a minimum at $1/2$ and tends to infinity at $0$ and $1$. The first order term is then directly proportional to the model variance but also inversely proportional to the ground truth certainty (i.e. how close $\hat{p}_1(\xt;\theta_0)$ is to 0 or 1). Thus, we see that this first order term matches the intuition behind epistemic uncertainty; if our predictions are uncertain but the true outcome is certain then we can expect to reduce our uncertainty and hence have high epistemic uncertainty.

\subsection{A Bootstrap Estimator}
\label{sec:bootstrap_estimator}

Theorem~\ref{thm:asymptotics} suggests that asymptotically correct estimation of the Fisher information and true model probabilities can thus be used to construct first order accurate estimators of mutual information. In light of this, we consider a bootstrap based estimator which we denote by $\I_b(\xt,\data)$.
\\
Estimation proceeds by first sampling weights $\xi$ from the symmetric Dirichlet with concentration parameter 1. Denote the MLE of $\theta$ over the bootstrapped dataset as $\thetab$ (for example, as obtained by minimizing cross-entropy loss). We denote the distribution of this estimate (induced by the randomness in the weights) given the data $\data$ as $p(\thetab|\data)$. Then,
\begin{align*}
&\I_b(\xt,\data):= \H(\E_{\thetab|\data}[\phatb]) - \E_{\thetab|\data}[\H(\phatb)].
\end{align*}
Algorithm~\ref{alg:bootstrap} makes this procedure explicit for clarity.
\begin{algorithm}[t]
  \caption{$\I_b(\xt,\data)$ estimation}
  \label{alg:bootstrap}
  \begin{algorithmic}
    \Statex {\bfseries Input:} Number of bootstrap replications $B$, training dataset $\data$
    \For{$b \leftarrow 1$ \textbf{to} $B$}
        \State \textbf{sample} $\xi \sim \mathrm{Dir}(1,...,1)$
        \State $\thetab \leftarrow \argmax_{\theta\in\Theta} \sum_{i=1}^n\xi_i\log(\hat{p}_{Y_i}(X_i;\theta))$
    \EndFor
    \Statex $\I_b(\xt,\data) = \H\left(\frac{1}{B}\sum_{b=1}^B\phatb\right) - \frac{1}{B}\sum_{b=1}^B\H\left(\phatb\right)$
    \Statex {\bfseries Output:} $\I_b(\xt,\data)$
  \end{algorithmic}
\end{algorithm}
\begin{theorem}
\label{theorem:bootstrap_correctness}
    Under the assumptions needed for Bernstein von-Mises to hold, for asymptotic normality of the MLE to hold and uniform integrability of the predicted probabilities (\ref{app:assumption:thm1:start}-\ref{app:assumption:thm2:final} in Appendix~\ref{sec:app:proofs}), we have
\begin{equation*}
  \frac{\I(\yt;\theta|\xt,\data)}{\I_b(\xt,\data)} \stackrel{P_{\theta_0}^n}{\rightarrow} 1 \qquad \forall\ \xt.
\end{equation*}
\end{theorem}

\textbf{Proof Sketch:} We proceed as in Theorem~\ref{thm:asymptotics} but invoke bootstrap limit laws (e.g. Theorem 10.16 of \citetay{kosorok_empirical_processes}) instead of Bernstein von-Mises to obtain an identical asymptotic expansion for our estimator. We provide the complete proof in Appendix~\ref{sec:app:proofs}. 

We note that our theorem also holds if we use the classic multinomial weights for bootstrap instead of Dirichlet ones\textemdash see Appendix~\ref{sec:app:proofs}. Theorem~\ref{theorem:bootstrap_correctness} proves the asymptotic validity of the bootstrap estimator and provides a frequentist path to studying the Bayesian MI. 


\subsection{Distinguishing Algorithmic Randomness from Statistical Uncertainty}
\label{sec:eu_decomp_theory}

Deep neural networks are generally trained with variants of stochastic gradient descent and a random parameter initialization. As such, the fitted model is random even with the training data fixed. Deep ensembles exploit this phenomenon, effectively capturing only the epistemic uncertainty arising from training stochasticity. The bootstrap estimator in Sec.~\ref{sec:bootstrap_estimator} provides the crucial conceptual link between the ad-hoc operational technique of deep ensembles and principled Bayesian inference. While standard Bayesian formulations cannot account for stochasticity in model training, in this section we show how the bootstrap estimator naturally decomposes into distinct terms for data variability and training stochasticity, providing a mechanism to isolate the latter. This decomposition will then allow us to study deep ensembles by directly comparing them to the training stochasticity component of a theoretically grounded estimator.

In more detail, the parameter estimate can be viewed as a function of the data and the random seed. Thus, for the remainder of this section, we use the notation $\phatb = \hat{p}(\datab,s)$ where $\datab$ denotes the bootstrapped dataset and $s$ denotes the seed used for training the DNN. Since $\thetab$ is deterministic given $(\datab,s)$, $\hat{p}(\datab,s)$ is a deterministic function. Normally, the selection of $s$ is completely independent of everything else and hence $(s|\datab,\data) \stackrel{d}{=} s$. 

With this notation in place, we propose the following decomposition of the bootstrap epistemic uncertainty estimator
\begin{align}
&\I_b(\xt,\data)  \nonumber\\
=& \H(\E_{(\datab,s)|\data}[\hat{p}(\datab,s)]) \pm \E_{\datab|\data}[\H(\E_{s}[\hat{p}(\datab,s)])] - \E_{(\datab,s)|\data}[\H(\hat{p}(\datab,s))] \nonumber\\
=& \underbrace{\text{\small$\H(\E_{\datab|\data}[\E_s[\hat{p}(\datab,s)]]) - \E_{\datab|\data}[\H(\E_{s}[\hat{p}(\datab,s)])]$}}_{=:\I^{\textrm{resampling}}_b} + \underbrace{\text{\small$\E_{\datab|\data}\left[\H(\E_{s}[\hat{p}(\datab,s)]) - \E_{s}[\H(\hat{p}(\datab,s))]\right]$}}_{=:\I^{\textrm{seeds}}_b}.
\label{eq:eu_decomp}
\end{align}
 $\I^{\textrm{resampling}}_b$ can be viewed as an estimator of the epistemic uncertainty arising due to the randomness in the data sampling process. Intuitively, this represents how much the model prediction would change if different training data had been sampled from the true distribution. On the other hand, $\I^{\textrm{seeds}}_b$ captures the amount of the entropy difference explained by the variation in the random seeds. That is, this term represents the amount of variation due to the algorithmic randomness. 
Thus, our proposed estimator can be decomposed into a `data sampling' component and a `training randomness' component.

Next, notice that the estimate of the mutual information from deep ensembles is
\begin{align}
\label{eq:deep_ensemble_eu}
    \I^{\textrm{deep ensemble}} := &\H(\E_{s}[\hat{p}(\data,s)]) - \E_{s}[\H(\hat{p}(\data,s))].
\end{align}
This is parallel to the term $\I^{\textrm{seeds}}_b$ above, but with the full data instead of a bootstrap subsample. Thus, the bootstrap estimator above can be viewed as approximately the MI from random seeds plus an additional term capturing variability in the dataset. In the experiments, we will see that the latter is often smaller than the former. In this way, $\I_b(\xt,\data) \approx  \I^{\textrm{deep ensemble}}.$
This link gives a new frequentist motivation for deep ensembles. More precisely, since Sec.~\ref{sec:bootstrap_estimator} proved that the bootstrap estimator is asymptotically correct for the true Bayesian target, the above relation indicates how deep ensembles theoretically succeed in capturing the Bayesian MI. 

\section{Experiments}
\label{sec:experiments}
Section~\ref{sec:eu_decomp_theory} gives us a way to measure contributions to epistemic uncertainty from sampling variability in the training data and from stochasticity in the training procedure; as well as a measure for the epistemic uncertainty in a deep ensemble. Our empirical investigations now address two questions\textemdash
\begin{enumerate}[leftmargin=*]
    \item Does training randomness ($\I^{\textrm{seeds}}_b$ in eq.~\ref{eq:eu_decomp}) contribute more to epistemic uncertainty than data sampling ($\I^{\textrm{resampling}}_b$ in eq.~\ref{eq:eu_decomp})?
    \item Do deep ensembles ($\I^{\textrm{deep ensemble}}$ in eq.~\ref{eq:deep_ensemble_eu}) capture the training randomness ($\I^{\textrm{seeds}}_b$)?
\end{enumerate}
\subsection{Experimental Setup}
We will use the same experimental setup across various deep learning classification tasks. We first train a deep ensemble of $B$ models on the entire training dataset. We use this to estimate $\I^{\textrm{deep ensemble}}$ for each test point by approximating expectations with empirical averages in eq.~\ref{eq:deep_ensemble_eu}. Next, we construct $B$ bootstrapped versions of the training dataset and train a deep ensemble of size $B$ on each. We use these deep ensembles to similarly estimate $\I^{\textrm{seeds}}_b$ and $\I^{\textrm{resampling}}_b$ for each test point. 
\subsection{Evaluation Metrics} 
\label{sec:eu_decomp_metrics}
To answer the first question, we look at the slope of the line of best fit ($m$ when $y=mx$) that regresses $\I^{\textrm{resampling}}_b$ on $\I^{\textrm{seeds}}_b$; we shall refer to this slope as $m_{\textrm{seeds}}^{\textrm{resampling}}$. Fig.~\ref{fig:eu_comparision_resampling} visualizes this line in red. This tells us how the magnitude of $\I^{\textrm{resampling}}_b$ compares to $\I^{\textrm{seeds}}_b$ on average.
For the second question, we again look at the slope of the line of best fit that regresses $\I^{\textrm{deep ensemble}}$ on $\I^{\textrm{seeds}}_b$ to compare the relative magnitudes of the two; we shall refer to this slope as $m_{\textrm{seeds}}^{\textrm{deep ensemble}}$. Fig.~\ref{fig:eu_comparision_deep_ensemble} \textit{(right)} visualizes this line in orange. We also look at the Spearman rank correlation between $\I^{\textrm{deep ensemble}}$ and $\I^{\textrm{seeds}}_b$ to evaluate if points with high relative epistemic uncertainty due to training stochasticity also exhibit high relative epistemic uncertainty when we use deep ensembles; we shall refer to this correlation as $\rho$. This is particularly insightful since performance on many downstream tasks (such as AUC-ROC in outlier detection tasks) often only depends on the ordering of test points rather the explicit magnitudes of the uncertainty estimates.
\subsection{Deep Learning Tasks}
To lend robustness to our empirical findings, we study performance across four tasks spanning various data domains (natural images, medical images, text), different dataset sizes ($1000$ classes with $>1$M training examples, $2$ classes with 1000 training examples) and different deep learning regimes (training from scratch, fine-tuning, transfer learning). For each task, we defer hyperparameter details and accuracy metrics to Appendix~\ref{app:sec:eu_decomp}.

\textbf{CIFAR-10:} We train a ResNet-18 \citep{resnets} from scratch on the CIFAR-10 dataset \citep{cifar10} which consists of 50,000 training examples spanning 10 classes of natural images. We use ensembles of size 25.

\textbf{ImageNet:} We train a ResNet-18 \citep{resnets} from scratch on the ImageNet dataset \citep{imagenet} which consists of 1.28M training examples spanning 1000 classes of natural images. Due to computational limitations, we use ensembles of size 10.

\textbf{ISIC Skin Lesion Dataset:} The ISIC (International Skin Imaging Collaboration) 2016 skin lesion challenge \citep{isic_dataset} consists of a dataset of images of skin lesions. There are only 900 training images spanning two classes (benign and malignant) in this dataset. For such small datasets, transfer learning \citep{transfer_learning} is a common approach wherein a pretrained model from a similar task is used to extract features for the task of interest. More precisely, we use a pretrained ResNet-50 model from \citetay{resnets} to extract 2048-dimensional feature vectors. We then train a two layer MLP from random initialization on top of these feature vectors to perform binary classification. We use ensembles of size 25.

\textbf{AG News Dataset:} The AG news topic classification dataset \citep{ag_news} consists of 120,000 news articles (titles and descriptions) spanning 4 categories. As is common in natural language processing (NLP), we finetune a pretrained model to perform classification. Specifically, we finetune the TinyBERT \citep{tinybert} model which has a transformer architecture (14.5M parameters) for classification by adding a linear layer (initialized randomly) on top of the transformer layers to perform classification. We use ensembles of size 25.
\subsection{Results}
\begin{figure}[t]
    \centering
    \includegraphics[width=1\linewidth]{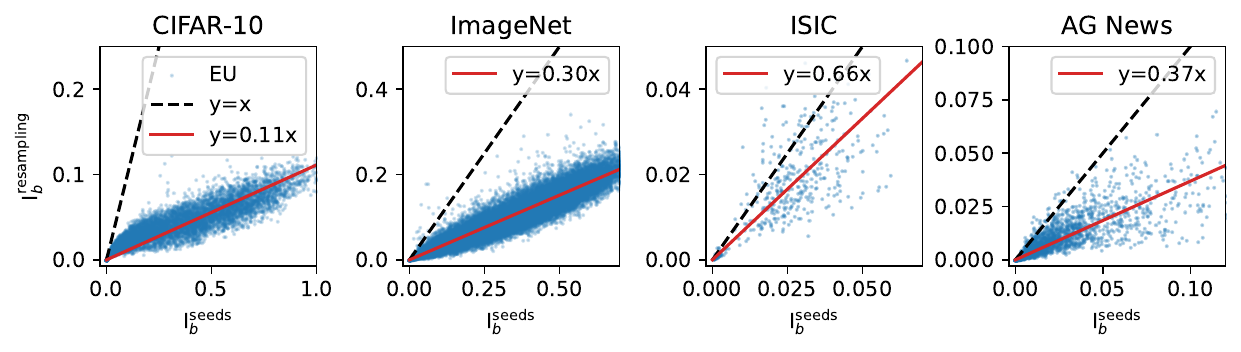}
    \caption{\textbf{Uncertainty Components of Bootstrap Estimator:} Scatter plot of epistemic uncertainty due to training stochasticity ($\I^{\textrm{seeds}}_b$) versus epistemic uncertainty due to data resampling ($\I^{\textrm{resampling}}_b$) along with line of best fit in red. See Sec. \ref{sec:eu_decomp_metrics} for details. }
    \label{fig:eu_comparision_resampling}
\end{figure}
\begin{figure}[t]
    \centering
    \includegraphics[width=1\linewidth]{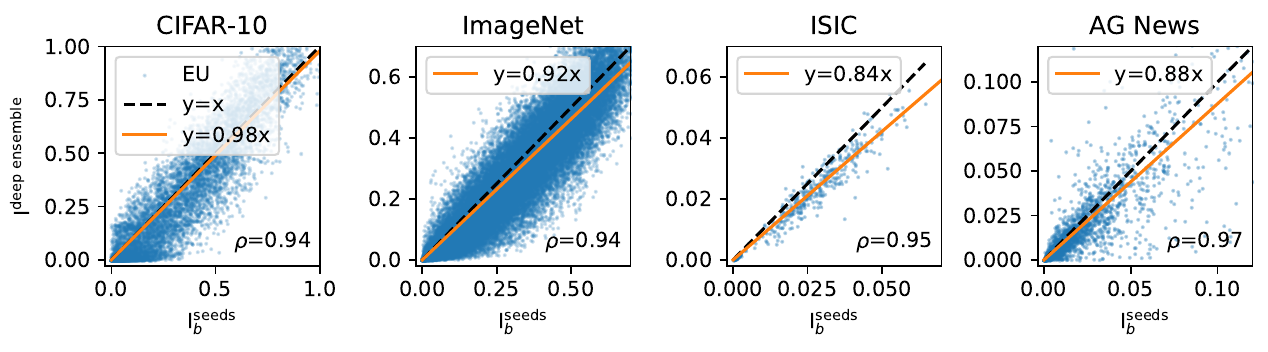}
    \caption{\textbf{Uncertainty due to Training Stochasticity:} Scatter plot of epistemic uncertainty due to training stochasticity ($\I^{\textrm{seeds}}_b$) versus deep ensemble estimate of epistemic uncertainty ($\I^{\textrm{deep ensemble}}$) along with line of best fit (orange) and spearman correlation ($\rho$). See Sec. \ref{sec:eu_decomp_metrics} for details. }
    \label{fig:eu_comparision_deep_ensemble}
\end{figure}
\begin{figure*}[t]
    \centering
    \begin{minipage}[b]{0.49\textwidth}
        \begin{center}
                \begin{sc}
                    {\footnotesize
                    \begin{tabular}{lccc}
                        \toprule
                       Dataset & $m_{\textrm{seeds}}^{\textrm{resampling}}$ & $m_{\textrm{seeds}}^{\textrm{deep ensemble}}$ & $\rho$ \\
                        \midrule
                        CIFAR-10  & 0.11 & 0.98 & 0.94 \\
                        ImageNet  & 0.30 & 0.92 & 0.94 \\
                        ISIC  & 0.66 & 0.84 & 0.95 \\
                        AG News  & 0.37 & 0.88 & 0.97 \\
                        \bottomrule
                    \end{tabular}}
                \end{sc}
        \end{center}
        \vspace{0.2in}
        \captionof{table}{Slope of line of best fit from $\I^{\textrm{resampling}}_b$ on $\I^{\textrm{seeds}}_b$ ($m_{\textrm{seeds}}^{\textrm{resampling}}$), from $\I^{\textrm{deep ensemble}}$ on $\I^{\textrm{seeds}}_b$ ($m_{\textrm{seeds}}^{\textrm{deep ensemble}}$) and spearman correlation between $\I^{\textrm{deep ensemble}}$ and  $\I^{\textrm{seeds}}_b$ ($\rho$). See Sec. \ref{sec:eu_decomp_metrics} for details.}
        \label{tab:eu_results}
    \end{minipage}%
    \hfill 
    \begin{minipage}[b]{0.49\textwidth}
    \centering
        \includegraphics[width=1\linewidth]{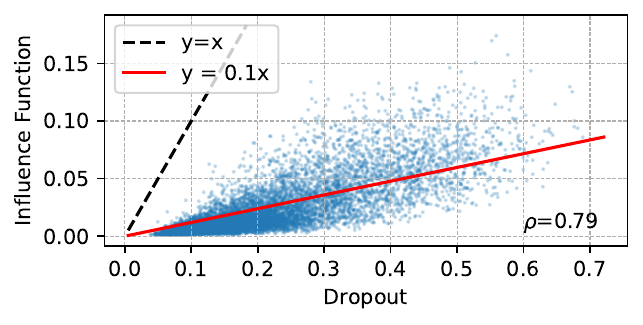}
        \caption{\textbf{Influence Function Approximation:} Estimates of epistemic uncertainty using Dropout, versus using influence functions to approximate predictions from bootstrapped datasets. See Sec.~\ref{sec:influence} for details.}
        \label{fig:influence_dropout}
    \end{minipage}
\end{figure*}

Fig.~\ref{fig:eu_comparision_resampling} visualizes scatter plots of $\I^{\textrm{seeds}}_b$ versus $\I^{\textrm{resampling}}_b$ along with the line of best fit (slope = $m_{\textrm{seeds}}^{\textrm{resampling}}$) with numerical values of the slopes in Table~\ref{tab:eu_results}. $\I^{\textrm{seeds}}_b$ always dominates over $\I^{\textrm{resampling}}_b$, with the latter being just 11\% of the former on average for CIFAR-10 for example. Therefore, training randomness contributes more to epistemic uncertainty than variability in data sampling. From a model development perspective, we believe that this suggests that if practitioners want to reduce their prediction uncertainty, it may be more fruitful to train a bigger deep ensemble to target the uncertainty due to stochastic training, than to collect more data to target the uncertainty due to variability in data sampling.

Fig.~\ref{fig:eu_comparision_deep_ensemble} visualizes scatter plots of $\I^{\textrm{seeds}}_b$ versus $\I^{\textrm{deep ensemble}}$ along with the line of best fit (slope = $m_{\textrm{seeds}}^{\textrm{deep ensemble}}$) with numerical values of the slopes and correlation in Table~\ref{tab:eu_results}. $\I^{\textrm{seeds}}_b$ and $\I^{\textrm{deep ensemble}}$ have similar magnitudes (around 90\% of each other) and correlate well ($\rho\approx0.95$). Therefore, deep ensembles do indeed capture the uncertainty due to training randomness. Together, these observations allow us to lend theoretical backing to deep ensembles since they successfully approximate the majority portion (training randomness) of an asymptotically correct estimator ($\I_b(\xt,\data)$) of the Bayesian epistemic uncertainty.

\section{Related Work}
\label{sec:related_works_au_eu}
\subsection{Uncertainty Decomposition} 
Mutual information as a measure of epistemic uncertainty has demonstrated success in active learning \citep{houlsby_bald, gal_deep_al}, out-of-distribution and misclassification detection tasks \citep{uq_decomp_nikita}. 
Beyond the specific information theoretic uncertainty decomposition considered in our work, a variety of other ways of quantifying epistemic uncertainty have also been proposed \citep{uq_decomp_senge, uq_decomp_kendall, uq_decomp_hofman, uq_decomp_lahou, uq_decomp_schweighofer, uq_decomp_tom} and MI does have drawbacks~\citep{mutual_info_critique}. See \citetay{aleatoric_epistemic_review} for a comprehensive review of the topic. Indeed, epistemic and aleatoric uncertainty are often not cleanly separable and take on different meanings in different contexts \citep{uq_decomp_critique}. We use MI for our investigations because of its widespread use, strong conceptual foundation and practical success; our work is not intended to address the optimality of MI as an epistemic uncertainty assessment.
\\
Similar to our work, \citetay{eu_decomp_lam} and \citetay{eu_decomp_aristide} also decompose the variance in model predictions for regression tasks into data variability and training randomness components. However, quite different to our contribution, \citetay{eu_decomp_lam} used this decomposition to study ways of theoretically eliminating training randomness using the NTK approximation \citep{NTK}, while \citetay{eu_decomp_aristide} used this decomposition to investigate the bias-variance tradeoff.

\subsection{Bayesian and Frequentist Equivalence}
\citetay{rubin_bayesian_bootstrap} considers bootstrap from a Bayesian lens, although this interpretation is hard to reconcile with traditional Bayesian inference since it considers priors over the data rather than the parameters. Closest to our theoretical results is the work by \citetay{michael_newton_wlb} which studied the use of bootstrap as a way of generating samples from a Bayesian posterior. Our work expands upon this idea by specifically leveraging this connection to connect frequentist and Bayesian notions of uncertainty. Moreover, we operationalize this idea in a modern context with DNNs and use it to study deep ensembles. Finally, \citetay{fong_martingale_posterior} also connects Bayesian parameter uncertainty to frequentist uncertainty arising due to lack of data. While their work conceptually shares connections to ours, it considers different practical applications.

\subsection{Deep Ensembles}
The uncertainty quantification capabilities of deep ensembles \citep{deep_ensembles} have been studied in depth \cite{random_seeds_uq_eval_ovadia, random_seeds_uq_eval_ashukha}. Broadly speaking, these studies observed that deep ensembles have well calibrated probabilistic forecasts both in and out-of-domain, and exhibit strong performance on downstream tasks such as out-of-distribution detection and misclassification detection. 
\\
Several reasons behind their success have been put forth \citep{ensembles_are_multimodal}, including Bayesian ones \citep{ensembles_are_bayesian}. These works primarily note that members of a deep ensemble capture different modes of the loss landscape and can be seen as high quality samples from the posterior. Our work is complementary to this perspective and provides a frequentist understanding of what uncertainties deep ensembles capture.

\subsection{Bootstrapping for Uncertainties in Deep Learning}
The idea of using bootstrap to estimate uncertainties in machine learning has been explored by several works in different contexts. \citetay{bootstrapping_edl_willem} investigates the suitability of evidential deep learning \citep{evidential_deep_learning} for uncertainty quantification by comparing them to the bootstrapped estimates of the model parameters, which they believe accurately capture uncertainty in model predictions due to randomness in the training data. \citetay{eu_decomp_willem} investigates variance based measures of uncertainty in a regression setting to also note that deep ensembles only capture uncertainty due to training variability which often dominates over uncertainty due to randomness in training data, similar to our findings in Sec.~\ref{sec:experiments}. Lastly, \citetay{eu_random_forests} have also considered approximating the posterior distribution by the trees in a random forest, which can be seen as bootstrap estimates, for the purposes of mutual information estimation as in our work.
\section{Discussion}
\subsection{Limitations} 
\label{sec:limitations}

Theorem~\ref{theorem:bootstrap_correctness} only proves the asymptotic validity of the bootstrap estimator which can weaken the theoretical link between deep ensembles and Bayesian inference in low data regimes. Furthermore, our theorems also rely on assumptions that can be unrealistic in practical settings. In particular, correct specification of the model and identifiability of parameters can certainly appear tenuous for DNNs. Finally, while we aimed to cover a wide range of deep learning settings in our empirical studies, we recognize that more evaluation can always be carried out to lend robustness to our conclusions. We do provide a few additional ablations against dataset size and choice of bootstrap weights in Appendix~\ref{app:sec:ablations} to lend further evidence towards the robustness of our qualitative conclusions.

\subsection{Future Directions} 
\label{sec:influence}
\subsubsection{Data Attribution Methods}
We identify an interesting link between the bootstrap estimator in Sec.~\ref{sec:bootstrap_estimator} and data attribution methods such as influence functions~\citep{hampel_influence_function, koh_influence_function, jackknife} and datamodels~\citep{datamodels, trak_madry} which aim to identify training data points most responsible for a model's predictions. These methods allow for estimation of predictions under perturbation of training data and can therefore be used to approximate the predictions of a model trained on bootstrapped data needed for our proposed measure. Any data attribution algorithm that yields a mapping from training data to model predictions can be leveraged to estimate the variance terms in~\eqref{eq:mi_expansion}, and hence give epistemic uncertainty estimates. We investigate the performance of such an MI estimate based on influence functions; see~Appendix~\ref{app:sec:data_attribution} for a similar experiment with datamodels and hyperparameter details.

We train a single convolutional network called the LeNet-5 on MNIST data  \citep{mnist}. Next, we calculate the influence functions for each data point as described in \citetay{koh_influence_function} for the model parameters in the last two fully connected layers. For each data point, we multiply the influence function by the change in the weighting under bootstrap and add all these products to the parameters to obtain an approximation for model parameters under a bootstrap reweighting. We obtain predictions from 100 such approximated models and use Algorithm~\ref{alg:bootstrap} to obtain MI estimates. To obtain a Bayesian estimate of MI, we use the Bayesian interpretation of Monte Carlo Dropout \cite{dropout} wherein Dropout is enabled at inference time and stochastic forward passes are performed to obtain samples from the Bayesian predictive distribution.

Fig.~\ref{fig:influence_dropout} shows the influence function approximation and Dropout estimates. While there is correlation, influence function estimates are an order of magnitude smaller. We hypothesize that this is because the influence function approximation only uses a single trained model and hence fails to capture the randomness due to training stochasticity. Nonetheless, our experiment serves as proof-of-concept for studying ``uncertainty attribution"\textemdash identifying data points or modalities responsible for model uncertainty\textemdash by estimating their influence function. Since the variance of an estimator is equal to variance of its influence function \citep{tsiatis_semiparametrics}, we can note that training data points with high influence on a test prediction drive up the variance of the influence function and hence lead to higher epistemic uncertainty. We hope to explore this more in a future work.

\subsubsection{Laplace Approximation}
We now note the connection between the Laplace approximation for Bayesian inference~\cite{mackay_bnns} and the asymptotic expansion for mutual information in Theorem~\ref{thm:asymptotics}. The Laplace approximation models the posterior $p(\theta|\data)$ as a Gaussian $\mathcal{N}(\hat{\theta}_{\textrm{MAP}}, \Sigma_{\textrm{Laplace}})$ centered at the maximum a posteriori (MAP) estimate $\hat{\theta}_{\textrm{MAP}}$, with precision matrix $\Sigma_{\textrm{Laplace}}^{-1}$ given by the negative Hessian of the log-posterior (the observed Fisher information plus the prior Hessian) evaluated at $\hat{\theta}_{\textrm{MAP}}$. Under suitable regularity conditions, the prior contribution is $O(1)$ while the normalized observed Fisher information converges to $\mathcal{I}(\theta_0)$, so that the Laplace posterior covariance $\Sigma_{\textrm{Laplace}}$ directly estimates the asymptotic covariance $n^{-1}\fisher$ in eq.~\ref{eq:mi_expansion}. Moreover, in finite samples where the empirical Fisher information may be ill-conditioned, the prior precision in $\Sigma_{\textrm{Laplace}}^{-1}$ provides natural regularization without altering the $O(n^{-1})$ asymptotics. 

Consequently, when using the Laplace approximation, eq.~\ref{eq:mi_expansion} can be used to directly provide an estimate of the epistemic uncertainty without explicit sampling from the posterior by evaluating $\hat{p}_k$ and its gradients at $\hat{\theta}_{\textrm{MAP}}$ and substituting $\Sigma_{\textrm{Laplace}}$ for $n^{-1}\fisher$. Empirically investigating the practicality of this idea and leveraging developments in computationally feasible Laplace approximations for deep learning~\cite{laplace_ggn, laplace_ggn_2, laplace_redux} to provide plug-in estimators of epistemic uncertainty is a promising future avenue.
\section{Conclusion}
Our paper aims to investigate the uncertainty quantification capabilities of deep ensembles from a frequentist perspective. Specifically, we study the epistemic uncertainty captured by deep ensembles, as measured by the mutual information between the model parameters and the data label. We derive an asymptotic expansion for this mutual information and use it to motivate the construction of an asymptotically correct bootstrap based estimator. Bayesian formulations cannot account for randomness in model parameter selection, while deep ensembles exclusively make use of this randomness. To bridge this gap, we make use of the bootstrap estimator\textemdash we show how the bootstrap estimate can be seen as an additive composition of the uncertainty from training randomness and uncertainty due to data sampling. We then conduct empirical studies which establish that training randomness dominates over data sampling, and that deep ensemble estimates of epistemic uncertainty closely capture the training randomness. Therefore, our investigations provide a theoretical frequentist justification to deep ensembles by noting that they approximate the dominant contribution to an asymptotically correct estimate of the true Bayesian epistemic uncertainty.

\section*{Acknowledgments}
The authors thank Willem Waegeman, Mira Jürgens and Sebastián Jiménez for their insightful comments and pointers to relevant literature. The authors would also like to thank Dan Kluger for his kind help and insights into the proofs. Finally, the authors would also like to thank the anonymous reviewers for their feedback which helped improve this paper.

\section*{Funding Transparency and Competing Interests Statement}
This research was, in part, funded by the Advanced Research Projects Agency for Health
(ARPA-H). The views and conclusions contained in this document are those of the authors
and should not be interpreted as representing the official policies, either expressed or implied,
of the United States Government. A.J. was also partly supported by the Michael Athans Fellowship. The authors declare no competing interests.

\newpage
\bibliography{references}
\bibliographystyle{plainnat}

\clearpage
\appendix
\thispagestyle{empty}

\newpage
\onecolumn
\section{Proofs}
\label{sec:app:proofs}
\subsection{Problem Setup and Notation:}
We refresh the problem setup and notation below. Consider a supervised learning $K$-class classification problem. Specifically,
\begin{enumerate}
    \item We assume that the training data $\data = \{(X_i,Y_i), i=1,...,n\}$ consists of feature-label pairs, where the features $X\in\mathcal{X}$ and the labels $Y\in\{1,...,K\}$.
    \item We wish to evaluate the epistemic uncertainty in the prediction for the label $\yt$ for the features $\xt$.
    \item We denote our predictions by $\phat$ where $\theta\in\Theta$ our model parameters. Further, denote by $\phatk$ the prediction probability assigned to the $k^{\textrm{th}}$ class. The model is specified such that $0<\phatk<1$ and $\sum_{k=1}^K\phatk=1$. 
    \item In the Bayesian case, we have a prior $p(\theta)$ over the parameters and $p(\theta|\data)$ is our posterior. Recall that the epistemic uncertainty measure in the Bayesian case is then
    \begin{align*}
    \I(\yt;\theta|\xt,\data) &= \H(\yt|\xt,\data) - \H(\yt|\theta,\xt,\data) \\
    &= \H(\E_{\theta\sim p(\theta|\data)}[\phat]) - \E_{\theta\sim p(\theta|\data)}[\H(\phat)].
    \end{align*}
    \item In the bootstrap case, we denote our estimate of $\theta$ over the bootstrapped dataset as $\thetab$. We further denote the distribution of this estimate (induced by the randomness in the choice of bootstrap weights) given the data $\data$ as $p(\thetab|\data)$. Recall that our proposed epistemic uncertainty measure is then
    \begin{equation*}
    \I_b(\xt,\data) = \H(\E_{\thetab\sim p_b(\thetab|\data)}[\phatb]) - \E_{\thetab\sim p_b(\thetab|\data)}[\H(\phatb)].
    \end{equation*}
    The bootstrap weights may be drawn from either a symmetric Dirichlet with concentration parameter 1 or a symmetric multinomial distribution.
\end{enumerate}
\subsection{Assumptions}
\begin{enumerate}
    \item \label{app:assumption:thm1:start} $\Theta\subset\mathbb{R}^p$ is open.
    \item There exists a $\theta_0\in\Theta$ such that $(X_i,Y_i)\stackrel{iid}{\sim}P_X(x)\hat{p}_y(x;\theta_0)=:P_{\theta_0}$.
    \item \label{app:assumption:first_derivative}Assume
    \begin{equation*}
        \psi_\theta(x,y) := \frac{\partial\ln(\hat{p}_y(x;\theta))}{\partial\theta}\Bigg|_{\theta=\theta} 
    \end{equation*} exists everywhere. Further assume $\E_{(x,y)\sim P_{\theta_0}}[\psi_{\theta_0}(x,y)]=0$.
    \item $\E_{(x,y)\sim P_{\theta_0}}[||\psi_{\theta_0}(x,y)||^2]<\infty$. Further, $\E_{(x,y)\sim P_{\theta_0}}[\psi_{\theta}(x,y)]$ is differentiable at $\theta_0$ with non singular derivative matrix $-\mathcal{I}(\theta_0)$. The model is sufficiently smooth such that $\E_{(x,y)\sim P_{\theta_0}}[\psi_{\theta_0}(x,y)\psi_{\theta_0}(x,y)^T] = \mathcal{I}(\theta_0)$.
    \item The prior $p(\theta)$ is absolutely continuous with respect to the Lebesgue measure in a neighborhood of $\theta_0$, with a continuous positive density at $\theta_0$.
    \item For every $\varepsilon>0$ there exists a sequence of tests $\phi_n$ such that 
    \begin{equation*}
        P^n_{\theta_0}(\phi_n(\data))\rightarrow0, \qquad \sup_{||\theta-\theta_0||>\varepsilon}P^n_{\theta}(1-\phi_n(\data))\rightarrow0.
    \end{equation*}
    \item \label{app:assumption:thm1:final} \label{app:assumption:bayes_ui}We further make the following assumption which implies uniform integrability to ensure that the first three moments of the posterior distribution converge
    \begin{equation*}
        \sup_n \E_{\theta\sim p(\theta|\data)}[|\sqrt{n}(\phatk-\hat{p}_k(\xt;\theta_0)|^{3+\delta}] < \infty \quad \forall \ k \quad \textrm{for some $\delta>0$}.
    \end{equation*}
    \item For any sequence $\{\theta_n\}\in\Theta$, $\E_{(x,y)\sim P_{\theta_0}}[\psi_{\theta_n}(x,y)]\rightarrow0$ implies $||\theta_n-\theta_0||\rightarrow0$.
    \item The class $\{\psi_\theta:\theta\in \Theta\}$ is strong Glivenko-Cantelli.
    \item For some $\eta>0$, the class $\mathcal{F}:=\{\psi_\theta:\theta\in\Theta, ||\theta-\theta_0||\leq\eta\}$ is Donsker and $\E_{(x,y)\sim P_{\theta_0}}[||\psi_{\theta}(x,y)-\psi_{\theta_0}(x,y)||^2]\rightarrow0$ as $||\theta-\theta_0||\rightarrow0$.
    \item Denote our parameter estimate over the complete dataset as $\hat{\theta}$. Then, $\frac{1}{n}\sum_{i=1}^n\psi_{\hat{\theta}}(X_i,Y_i)=o_p(n^{-1/2})$. Likewise, for our bootstrap estimate, $\sum_{i=1}^n\xi_i\psi_{\thetab}(X_i,Y_i)=o_p(n^{-1/2})$ where $\xi_i,i=1,...,n$ are the bootstrap weights.
    \item \label{app:assumption:thm2:final} We again make the following assumption which implies uniform integrability to ensure that the first three moments of the bootstrap distributions converge:
    \begin{equation*}
        \sup_n \E_{\thetab\sim p(\thetab|\data)}[(\sqrt{n}|\phatkb-\hat{p}_k(\xt;\theta_0)|^{3+\delta}] < \infty \quad \forall \ k \quad \textrm{for some $\delta>0$}.
    \end{equation*}
\end{enumerate}
\subsection{Theorem 1}
Under the assumptions \ref{app:assumption:thm1:start}-\ref{app:assumption:thm1:final}, we have the asymptotic expansion
\begin{equation*}
    \I(\yt;\theta|\xt,\data) = \frac{1}{2n}\sum_{k=1}^K \frac{\sigma_k^2}{\hat{p}_k(\xt;\theta_0)} +o_p(n^{-1}) \qquad \forall \ \xt,
\end{equation*}
where $\sigma_k^2$ is
\begin{equation*}
    \sigma_k^2 = \left[\frac{\partial\hat{p}(\xt;\theta_0)}{\partial\theta}^{T}\fisher\frac{\partial\hat{p}(\xt;\theta_0)}{\partial\theta}\right]_{k,k},
\end{equation*}
for $\fisher$ being the inverse Fisher information of the model at $\theta_0$.

\subsubsection{Proof}

Assume that we have some prior $p(\theta)$ over our parameters and denote our posterior by $p(\theta|\data)$. Note that our epistemic uncertainty measure is then
\begin{align*}
    \I(\yt;\theta|\xt,\data) &= \H(\yt|\xt,\data) - \H(\yt|\theta,\xt,\data) \\
    &= \H(\E_{\theta\sim p(\theta|\data)}[\phat]) - \E_{\theta\sim p(\theta|\data)}[\H(\phat)].
\end{align*}
Our proof proceeds in two parts, we first use Taylor's theorem to find an expansion of the epistemic uncertainty and then we find the asymptotic limit.

Note that $\H$ is infinitely differentiable with first three derivatives given by
\begin{align*}
    \frac{\partial \H(\hat{p})}{\partial \hat{p}_i} &= -1 -\log(\hat{p}_i), \\
    \frac{\partial^2 \H(\hat{p})}{\partial \hat{p}_i\partial \hat{p}_j} &= \1\{i=j\}\left(\frac{-1}{\hat{p}_i}\right), \\
    \frac{\partial^3 \H(\hat{p})}{\partial \hat{p}_i\partial \hat{p}_j \partial\hat{p}_k} &= \1\{i=j=k\}\left(\frac{1}{\hat{p}_i^2}\right).
\end{align*}
Taylor expanding $\H(\phat)$ around $\E_{\theta\sim p(\theta|\data)}[\phat]$ and using the Lagrange form of the remainder yields
\begin{align*}
    \H(\hat{p}(\xt;\theta)) =& \H(\E_{\theta\sim p(\theta|\data)}[\hat{p}(\xt;\theta)]) \\
    &+ \sum_{k=1}^K (\hat{p}_k(\xt;\theta) - \E_{\theta\sim p(\theta|\data)}[\hat{p}_k(\xt;\theta)])(-1-\log(\E_{\theta\sim p(\theta|\data)}[\hat{p}_k(\xt;\theta)]) \\
    &+ \frac{1}{2} \sum_{k=1}^K (\hat{p}_k(\xt;\theta) - \E_{\theta\sim p(\theta|\data)}[\hat{p}_k(\xt;\theta)])^2\left(\frac{-1}{\E_{\theta\sim p(\theta|\data)}[\hat{p}_k(\xt;\theta)]}\right) \\ 
    &+ \frac{1}{6}\sum_{k=1}^K (\hat{p}_k(\xt;\theta) - \E_{\theta\sim p(\theta|\data)}[\hat{p}_k(\xt;\theta)])^3\left(\frac{1}{\hat{p}_k'^2}\right),
\end{align*}
where $\hat{p}_k'$ is some value between $\hat{p}_k(\xt;\theta)$ and $\E_{\theta\sim p(\theta|\data)}[\hat{p}_k(\xt;\theta)]$ for $k=1,...,K$.

Substituting this back into the expression for epistemic uncertainty yields
\begin{align*}
    & \I(\yt;\theta|\xt,\data) \\
    &= \H(\E_{\theta\sim p(\theta|\data)}[\hat{p}(\xt;\theta)]) \\
    &- \E_{\theta\sim p(\theta|\data)}[\H(\E_{\theta\sim p(\theta|\data)}[\hat{p}(\xt;\theta)])] \\
    &- \E_{\theta\sim p(\theta|\data)}\left[\sum_{k=1}^K (\hat{p}_k(\xt;\theta) - \E_{\theta\sim p(\theta|\data)}[\hat{p}_k(\xt;\theta)])(-1-\log(\E_{\theta\sim p(\theta|\data)}[\hat{p}_k(\xt;\theta)])\right] \\
    &- \frac{1}{2}\E_{\theta\sim p(\theta|\data)}\left[\sum_{k=1}^K (\hat{p}_k(\xt;\theta) - \E_{\theta\sim p(\theta|\data)}[\hat{p}_k(\xt;\theta)])^2\left(\frac{-1}{\E_{\theta\sim p(\theta|\data)}[\hat{p}_k(\xt;\theta)]}\right)\right] \\
    &- \frac{1}{6}\E_{\theta\sim p(\theta|\data)}\left[\sum_{k=1}^K (\hat{p}_k(\xt;\theta) - \E_{\theta\sim p(\theta|\data)}[\hat{p}_k(\xt;\theta)])^3\left(\frac{1}{\hat{p}_k'^2}\right)\right]. \\
\end{align*}
Note that the first and second term cancel, and the third term is equal to zero.
Thus,
\begin{align*}
    \I(\yt;\theta|\xt,\data) =& \frac{1}{2} \sum_{k=1}^K \left(\frac{\E_{\theta\sim p(\theta|\data)}\left[(\hat{p}_k(\xt;\theta) - \E_{\theta\sim p(\theta|\data)}[\hat{p}_k(\xt;\theta)])^2\right]}{\E_{\theta\sim p(\theta|\data)}[\hat{p}_k(\xt;\theta)]}\right) \\
    &- \frac{1}{6} \sum_{k=1}^K \left(\E_{\theta\sim p(\theta|\data)}\left[\frac{(\hat{p}_k(\xt;\theta) - \E_{\theta\sim p(\theta|\data)}[\hat{p}_k(\xt;\theta)])^3}{\hat{p_k'^2}}\right]\right). \\
\end{align*}
Note that the first term above is the variance of the probability predictions divided by the mean prediction under the posterior distribution, i.e.
\begin{align*}
    \I(\yt;\theta|\xt,\data) =& \frac{1}{2} \sum_{k=1}^K \left(\frac{\var_{\theta\sim p(\theta|\data)}\left[\phatk\right]}{\E_{\theta\sim p(\theta|\data)}[\hat{p}_k(\xt;\theta)]}\right) \\
    &- \frac{1}{6} \sum_{k=1}^K \left(\E_{\theta\sim p(\theta|\data)}\left[\frac{(\hat{p}_k(\xt;\theta) - \E_{\theta\sim p(\theta|\data)}[\hat{p}_k(\xt;\theta)])^3}{\hat{p_k'^2}}\right]\right). \\
\end{align*}
Next, we shall evaluate the limit. \\
Under assumptions ~\ref{app:assumption:thm1:start}-~\ref{app:assumption:thm1:final}, we can apply the Bernstein-von Mises theorem (Theorem 10.1 of \citetay{vdv}). Note that Bernstein von-Mises implies convergence in total variation distance which directly implies convergence in distribution. Hence, we have
\begin{equation*}
    \sqrt{n}\left(\theta-\theta_0\right)\big|\data \stackrel{d}{\rightarrow} \cN(0,\fisher) \qquad \textrm{in $P_{\theta_0}^n$-probability}.
\end{equation*}
In particular, the above is short-hand for
\begin{equation*}
    \sup_{h\in BL_1}\left|\E_{\theta\sim p(\theta|\data)}\left[h\left(\sqrt{n}(\theta-\theta_0)\right)\right]-\E_{Z\sim\cN(0,\fisher)}[h(Z)]\right| \stackrel{P_{\theta_0}^n}{\rightarrow} 0,
\end{equation*}
where $BL_1$ is the space of functions $h:\mathbb{R}^p\mapsto\mathbb{R}$ with Lipschitz norm bounded by 1. Recall that if both expectations on the left hand side were unconditional, limit of the left hand side going to 0 is the standard definition of convergence in distribution. See Sec. 23.2.1 of \citetay{vdv} for further discussion. 

We shall henceforth continue with the ``$(.) \stackrel{d}{\rightarrow}$ (.) \quad in $P_{\theta_0}^n$ probability'' notation for ease of exposition.

Note that \ref{app:assumption:first_derivative} allows us to apply the functional delta method (see chapter 20 of \citetay{vdv}) to evaluate the limit of the function $\theta\mapsto\phatk$ under the posterior. This yields,
\begin{equation}
    \sqrt{n}\left(\phatk - \hat{p}_k(\xt;\theta_0)\right)\big|\data \stackrel{d}{\rightarrow} \cN(0, \sigma^2_k) \qquad \textrm{in $P_{\theta_0}^n$-probability},
    \label{app:eq:bvm_delta}
\end{equation}
where $\sigma^2_k = \left[\frac{\partial\hat{p}(\xt;\theta_0)}{\partial\theta}^{T}\fisher\frac{\partial\hat{p}(\xt;\theta_0)}{\partial\theta}\right]_{k,k}$. Finally, \ref{app:assumption:bayes_ui} implies uniform integrability allowing us to conclude that the limit of the moment is the moment of the limiting distribution:
\begin{equation*}
    \E_{\theta\sim p(\theta|\data)} [\hat{p}_k(\xt;\theta)] \stackrel{P_{\theta_0}^n}{\rightarrow}  \hat{p}_k(\xt;\theta_0),
\end{equation*}
\begin{equation*}
    \var_{\theta\sim p(\theta|\data)}\left[\sqrt{n}\phatk\right] \stackrel{P_{\theta_0}^n}{\rightarrow} \sigma_k^2 .
\end{equation*}

Next, we shall consider the remainder term from the Taylor expansion. Note that $\hat{p}_k' \in [\hat{p}_k(\xt;\theta),\E_{\theta\sim p(\theta|\data)}[\hat{p}_k(\xt;\theta)]]$. By assumption, $0<\phatk < 1$ and hence $1/\hat{p}_k'^2 = O_p(1)$. Furthermore, we also have from equation~\ref{app:eq:bvm_delta} that $\left(\phatk - \hat{p}_k(\xt;\theta_0)\right)\big|\data = O_p\left(\frac{1}{\sqrt{n}}\right)$. Therefore,
\begin{equation*}
    n\frac{(\hat{p}_k(\xt;\theta) - \E_{\theta\sim p(\theta|\data)}[\hat{p}_k(\xt;\theta)])^3}{\hat{p_k'^2}} = nO_p\left(\frac{1}{n^{3/2}}\right)O_p(1) = O_p\left(\frac{1}{\sqrt{n}}\right).
\end{equation*}
Again, by the uniform integrability assumption
\begin{equation*}
    \E_{\theta\sim p(\theta|\data)}\left[n\frac{(\hat{p}_k(\xt;\theta) - \E_{\theta\sim p(\theta|\data)}[\hat{p}_k(\xt;\theta)])^3}{\hat{p_k'^2}}\right] \stackrel{P_{\theta_0}^n}{\rightarrow} 0 .
\end{equation*}
Therefore,
\begin{equation*}
n\I(\yt;\theta|\xt,\data) \stackrel{P_{\theta_0}^n}{\rightarrow} \frac{1}{2}\sum_{k=1}^K \frac{\sigma_k^2}{\hat{p}_k(\xt;\theta_0)}.
\end{equation*}

\subsection{Theorem 2} 
Under the assumptions \ref{app:assumption:thm1:start}-\ref{app:assumption:thm2:final}, 
\begin{equation*}
    \frac{\I(\yt;\theta|\xt,\data)}{\I_b(\xt,\data)} \stackrel{P_{\theta_0}^n}{\rightarrow} 1 \qquad \forall \ \xt,
\end{equation*}
In particular, note that $\I_b$ may be calculated using either symmetric Dirichlet weights with concentration parameter 1, or multinomial weights.

\subsubsection{Proof}

Recall that our proposed epistemic uncertainty measure for the bootstrap case is
\begin{equation*}
    \I_b(\xt,\data) = \H(\E_{\thetab\sim p_b(\thetab|\data)}[\phatb]) - \E_{\thetab\sim p_b(\thetab|\data)}[\H(\phatb)] .
\end{equation*}
We can proceed exactly as in the Bayesian case to obtain 
\begin{align*}
    \I_b(\xt,\data) =& \frac{1}{2} \sum_{k=1}^K \left(\frac{\var_{\thetab\sim p_b(\thetab|\data)}\left[\phatkb\right]}{\E_{\thetab\sim p_b(\thetab|\data)}[\hat{p}_k(\xt;\thetab)]}\right) \\
    &- \frac{1}{6} \sum_{k=1}^K \left(\E_{\thetab\sim p_b(\thetab|\data)}\left[\frac{(\hat{p}_k(\xt;\thetab) - \E_{\thetab\sim p_b(\thetab|\data)}[\hat{p}_k(\xt;\thetab)])^3}{\hat{p_k'^2}}\right]\right), \\
\end{align*}
where $\hat{p}_k'$ is now some value between $\hat{p}_k(\xt;\thetab)$ and $\E_{\thetab\sim p_b(\thetab|\data)}[\hat{p}_k(\xt;\thetab)]$ for $k=1,...,K$.
Next, from Theorem 10.16 of \citetay{kosorok_empirical_processes}, we have
\begin{equation*}
    \sqrt{n}\left(\thetab-\hat{\theta}\big|\data\right) \stackrel{d}{\rightarrow} \cN(0,\fisher) \qquad \textrm{in $P_{\theta_0}^n$-probability}
\end{equation*}
and
\begin{equation*}
    \sqrt{n}\left(\hat{\theta}-\theta_0\right) \stackrel{d}{\rightarrow} \cN(0,\fisher),
\end{equation*}
where $\hat{\theta}$ is the MLE over the complete data. Notably, both the statements above hold for either multinomial or Dirichlet weights.

We can now apply the delta method for bootstrap (\citet{vdv} Theorem 23.9) to this and proceed exactly as in the Bayesian case to conclude
\begin{equation*}
n\I_b(\xt,\data) \stackrel{P_{\theta_0}^n}{\rightarrow} \frac{1}{2}\sum_{k=1}^K \frac{\sigma_k^2}{\hat{p}_k(\xt;\theta_0)}.
\end{equation*}
Combining the Bayesian and bootstrap results by the continuous mapping theorem,
\begin{equation*}
    \frac{n\I(\yt;\theta|\xt,\data)}{n\I_b(\xt,\data)} \stackrel{P_{\theta_0}^n}{\rightarrow} 1.
\end{equation*}
\newpage
\section{Additional Experiments and Experimental Details}
All experiments were conducted on an Ubuntu 22.04.5 Linux cluster using 2 NVIDIA V100 Tensor Core GPUs. All experiments needed no more than 32GB of GPU memory and 40GB of CPU RAM. Training the ImageNet models for a single seed took roughly 8 hours per seed, CIFAR-10 took about 30 minutes per seed, AGNews took less than 20 minutes per seed and ISIC took less than a minute per seed. Total time to reproduce results in the paper is no more than 30 days of computation with resources listed above and total computation time used by the research project is no more than 100 days.
\label{sec:app:experiments}
\subsection{Epistemic Uncertainty Decomposition}
\label{app:sec:eu_decomp}
\subsubsection{CIFAR-10}
\textbf{Experimental Details:} We use the CIFAR-10 dataset \citep{cifar10} (publicly available granting all rights to users) with the original training and test set folds.
We use a standard ResNet-18 \citep{resnets} architecture but modify the first convolutional layer to have a kernel size of 3, stride of 1, padding of 1 to allow the network to work with 32
$\times$32 size images. Further, we modify the last fully connected layer to only have 10 output neurons for a 10 class classification task. We use cross entropy loss and Adam \citep{adam} as an optimizer with learning rate of 0.001 and betas of 0.9 and 0.999. We use a batch size of 128 and train for 15 epochs. The rest of the experiment proceeds as described in the main text.

\textbf{Accuracy Metrics:} Models within the deep ensemble achieve an accuracy of 0.83 on average, whereas the bootstrap models achieve an accuracy of 0.80 on average. The deep ensemble itself (i.e. averaged model predictions) achieves an accuracy of 0.89. Deep ensembles over the bootstrapped datasets (i.e. model predictions averaged over the random seeds only) achieves an accuracy of 0.88 on average across the bootstrapped datasets.

\subsubsection{ImageNet}
\textbf{Experimental Details:} We use the ImageNet dataset \citep{imagenet} (available for non-commercial use subject to terms of usage listed at \url{https://www.image-net.org/}) with the original training and test set folds. For computational ease, we use a downsampled version of the images downsampled to $32\times32$ pixels as in \citetay{imagenet_downsampled}.
We use a standard ResNet-18 \citep{resnets} architecture but modify the first convolutional layer to have a kernel size of 3, stride of 1, padding of 1 to allow the network to work with 32
$\times$ 32 size images. We use cross entropy loss and SGD as an optimizer with learning rate of 0.1, momentum of 0.9 and weight decay of 0.0001. We use a batch size of 128 and train for 40 epochs. We use the one cycle learning rate policy from \citetay{lr_schedule}. The rest of the experiment proceeds as described in the main text.

\textbf{Accuracy Metrics:} Models within the deep ensemble achieve top-1 and top-5 accuracy of 0.55 and 0.79 respectively on average, whereas the bootstrap models achieve top-1 and top-5 accuracies of 0.51 and 0.75 respectively on average. The deep ensemble itself (i.e. averaged model predictions) achieves top-1 and top-5 accuracy of 0.59 and 0.81 respectively. Deep ensembles over the bootstrapped datasets (i.e. model predictions averaged over the random seeds only) achieves top-1 and top-5 accuracy of 0.55 and 0.78 on average across the bootstrapped datasets.

\subsubsection{ISIC Dataset}
\textbf{Experimental Details:} 
We use the ISIC Skin Lesion challenge (2016) dataset \citep{isic_dataset} (publicly available for non-commerical research and educational use) with the original training and test set folds. We pass all images through the trained ResNet-50 from \citetay{resnets} and extract the activations from the last layer to obtain 2048-dimensional feature vectors. We then train a two layer neural network from scratch on top of these feature vectors with 512 hidden units, ReLU activation and Dropout with a rate of 0.5. We use Adam \citep{adam} as an optimizer with a learning rate of 0.001 and betas of 0.9 and 0.999. We use a batch size of 32 and train for 15 epochs minimizing cross entropy loss weighted by prevalence of the ground truth class in the training data to account for class imbalance. The rest of the experiment proceeds as described in the main text.

\textbf{Accuracy Metrics:} Models within the deep ensemble achieve an AUC-ROC of 0.81 on average, whereas the bootstrap models achieve an accuracy of 0.79 on average. The deep ensemble itself (i.e. averaged model predictions) achieves an accuracy of 0.81. Deep ensembles over the bootstrapped datasets (i.e. model predictions averaged over the random seeds only) achieves an accuracy of 0.80 on average across the bootstrapped datasets.

\subsubsection{AG News}
\textbf{Experimental Details:} We use the AG News dataset \citep{ag_news} (publicly available for non-commerical research and educational use) with the original training and test set folds. We finetune the TinyBERT model from \citetay{tinybert}. We add a linear layer on top of the final 312 dimensional output from TinyBERT to perform 4 class classification (parameters initialized randomly). We use Adam \citep{adam} as an optimizer with a learning rate of $2e-5$ and betas of 0.9 and 0.999. We also use a linear learning rate schedule with warmup as provided by the transformers \citep{transformers_library} library. We train for 3 epochs with a batch size of 32. We tokenize each example up to 128 tokens, padding or truncating as necessary. The rest of the experiment proceeds as described in the main text.

\textbf{Accuracy Metrics:} Models within the deep ensemble achieve an accuracy of 0.94 on average, whereas the bootstrap models achieve an accuracy of 0.93 on average. The deep ensemble itself (i.e. averaged model predictions) achieves an accuracy of 0.94. Deep ensembles over the bootstrapped datasets (i.e. model predictions averaged over the random seeds only) achieves an accuracy of 0.93 on average across the bootstrapped datasets.

\subsection{Data Attribution}
\label{app:sec:data_attribution}
\textbf{Experimental Details:} We use the MNIST dataset \citep{mnist} (available under a CC BY-SA 3.0 license). For training, we randomly select 600 images from each class to form the training dataset and keep the test dataset the same as the original.

We use a two layer convolutional neural network with 3 fully connected layers. We use Adam \citep{adam} as an optimizer with a learning rate of 0.001, betas of 0.9 and 0.999, weight decay of 0.0001, and a batch size of 32. We train for 10 epochs.

To calculate the influence function, we freeze all but the final two fully connected layers and calculate the hessian and gradients only with respect to the parameters in these layers. Specifically, as in \citetay{koh_influence_function}, the change in model parameters (for the final two layers) for an $\epsilon$ increase in the weighting of the $i^{th}$ training data point is $-H_{\hat{\theta}}^{-1}\nabla_\theta \log(\hat{p}_{Y_i}(X_i;\hat{\theta}))$
where $\hat{\theta}$ is the MLE on the original (unweighted) dataset, $H_{\hat{\theta}} = \frac{1}{n} \sum_{i=1}^n \nabla^2_\theta \log(\hat{p}_{Y_i}(X_i;\hat{\theta}))$ is the hessian and $\nabla_\theta$ denotes derivative with respect to $\theta$. We use multinomial bootstrap weights. For practical calculation, we add the identity matrix scaled by $10^{-5}$ to the Hessian to ensure invertibility in cases with non-negative definite hessian\textemdash see \citetay{koh_influence_function} for discussion.

For the Dropout estimate, we add Dropout layers with probability 0.5 between the fully connected layers and train with the same hyperparameters above. 

The rest of the experiment proceeds as in the main text.

\subsubsection{Datamodels Experiment}
\begin{figure}
    \centering
    \includegraphics[width=0.75\linewidth]{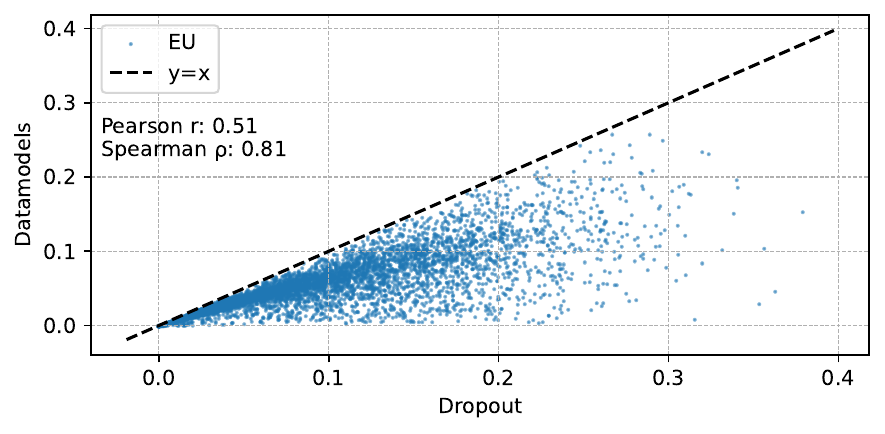}
    \caption{\textbf{Datamodels Approximation:} Datamodel approximation to our measure versus mutual information estimates obtained using Dropout on CIFAR-10 test points. See text for details.}
    \label{fig:app:datamodels}
\end{figure}
We now also investigate the use of datamodels \citep{datamodels} as a means of approximating our measure.

\textbf{Experimental Details:} We use the CIFAR-10 dataset \citep{cifar10} and ResNet-9 architecture as in \citetay{datamodels}. Each test point is associated with a datamodel which consists of a weight vector of length 50001\textemdash the number of training data points and a bias term. The datamodel is such that it predicts the model output for that test data point given the weights assigned to the training data as the dot product between the training data weights and the datamodel weights. The model output considered in \citetay{datamodels} is unfortunately not the entire probability vector, but simply the difference between the correct logit and the highest incorrect logit. For our experiment, we use the datamodels for CIFAR-10 provided as part of the paper\textemdash see \citetay{datamodels} for details. We then draw 100 sets of Dirichlet weights and obtain estimates of model outputs using the datamodel. We then pass the model output through the sigmoid function to obtain a probability and use this for calculation of the EU. Note that this is equivalent to treating the classification task as binary between the correct class vs the highest incorrect class. While this is not equivalent to the correct calculation of EU over the 10 class probability vector, it is the best we can do with the existing datamodel.

For the dropout estimate, we train a ResNet-9 with Dropout layers added with probability 0.5 with the exact same hyperparameters used to create the datamodel.

\textbf{Results:} Fig.~\ref{fig:app:datamodels} shows that the datamodels approximation to our measure and the Dropout epistemic uncertainty scores have a high rank correlation and moderate Pearson correlation. The datamodel estimates are also consistently lower than the Dropout estimates. This may be because of the reduction to a binary classification task in the Datamodels approximation versus the full 10 class classification in the Dropout calculation. Nonetheless, this is a useful proof of concept experiment which opens the door to using data attribution methods for a more nuanced understanding of sources of uncertainty.

\subsection{Ablations}
\label{app:sec:ablations}
\subsubsection{Choice of Bootstrap Weights}
The experimental results in Sec.~\ref{sec:experiments} were conducted with Dirichlet weights for the bootstrap instead of the more common multinomial weights. As discussed in Sec.~\ref{sec:background}, we hypothesize that these weights are more suitable since it allows the neural network to still see all the training instances and hence capture the training data better than multinomial weights which omit training points entirely. We now ablate against this choice of weights. Specifically, we now perform bootstrapping by drawing bootstrap datasets by performing i.i.d. draws from the empirical distribution. This is the same as using bootstrap weights drawn from the multinomial distribution with event probabilities all being symmetrically $1/n$ where $n$ is the size of the dataset (assuming no repeated datapoints).

\begin{table}[htbp]
\centering
\renewcommand{\arraystretch}{1.5}
\begin{tabular}{llccc}
\toprule
 & \textbf{Dataset} & ${m_\textrm{seeds}^\textrm{resampling}}$ & ${m_\textrm{seeds}^\textrm{deep ensemble}}$ & $\rho$ \\
\midrule
\multirow{3}{*}{\rotatebox{90}{Multinomial}} 
 & CIFAR-10 & 0.24 & 0.82 & 0.89 \\
 & AG News  & 0.47 & 0.84 & 0.96 \\
 & ISIC     & 0.61 & 0.77 & 0.88 \\
\midrule
\multirow{3}{*}{\rotatebox{90}{Dirichlet}} 
 & CIFAR-10 & 0.11 & 0.98 & 0.94 \\
 & AG News  & 0.37 & 0.88 & 0.97 \\
 & ISIC     & 0.66 & 0.84 & 0.95 \\
\bottomrule
\end{tabular}
\vspace{10pt}
\caption{Comparison of slopes of line of best fit and correlations when using multinomial versus dirichlet weights across datasets.}
\label{app:tab:multinomial_ablation}
\end{table}

Table~\ref{app:tab:multinomial_ablation} reports the metrics defined in Sec.~\ref{sec:eu_decomp_metrics} for the two different choices of weights. Since multinomial weights omit datapoints entirely from training, they tend to lead to more significant variation when bootstrapping and hence the resampling component becomes more significant while the correlation between the deep ensemble estimate and seeds component worsens. Yet, qualitatively, the seeds component still dominates and correlates well with the deep ensemble estimate.

\subsubsection{Dataset Size}
We additionally ablate against the size of the dataset to investigate whether our conclusions still hold in low data regimes where the resampling component may be more dominant. On the CIFAR-10 dataset, we subsample to 5\% and 25\% of the full training data and then repeat our experiments. Table~\ref{app:tab:datasize_ablation} reports values of the metrics defined in Sec.~\ref{sec:eu_decomp_metrics}.

\begin{table}[htbp]
\centering
\renewcommand{\arraystretch}{1.5} 
\begin{tabular}{lccc}
\toprule
 & \multicolumn{3}{c}{\textbf{Dataset Size}} \\
\cmidrule(lr){2-4}
 & 5\% & 25\% & 100\% \\
\midrule
$m_\text{seeds}^\text{resampling}$   & 0.21 & 0.22 & 0.11 \\
$m_\text{seeds}^\text{deep ensemble}$ & 1.05 & 1.26 & 0.98 \\
$\rho$                     & 0.79 & 0.87 & 0.94 \\
\bottomrule
\end{tabular}

\vspace{10pt} 
\caption{Comparison of slopes of line of best fit and correlation when using subsampled versions of the CIFAR-10 dataset.}
\label{app:tab:datasize_ablation}
\end{table}

The contribution to uncertainty from resampling the data does grow slightly as expected, but continues to be clearly dominated by the seeds contribution (even at just 5\% of the original data). Further, while the correlation between the deep ensemble estimate and the seeds component worsens, it still stays quite high. Our conclusions thus seem robust to the dataset size.


\newpage
\section*{NeurIPS Paper Checklist}

\begin{enumerate}

\item {\bf Claims}
    \item[] Question: Do the main claims made in the abstract and introduction accurately reflect the paper's contributions and scope?
    \item[] Answer: \answerYes{}{} 
    \item[] Justification: Sec. 3 presents all theoretical results and Sec. 4 combines with experimental findings to prove main claims.

\item {\bf Limitations}
    \item[] Question: Does the paper discuss the limitations of the work performed by the authors?
    \item[] Answer: \answerYes{}{} 
    \item[] Justification: See Sec. 6

\item {\bf Theory assumptions and proofs}
    \item[] Question: For each theoretical result, does the paper provide the full set of assumptions and a complete (and correct) proof?
    \item[] Answer: \answerYes{}
    \item[] Justification: The main assumptions and proof sketch are given when presenting the theorems in Sec. 3 itself, with formal mathematical statements for all assumptions and rigorous proofs in the Supplement Sec. A. 

    \item {\bf Experimental result reproducibility}
    \item[] Question: Does the paper fully disclose all the information needed to reproduce the main experimental results of the paper to the extent that it affects the main claims and/or conclusions of the paper (regardless of whether the code and data are provided or not)?
    \item[] Answer: \answerYes{} 
    \item[] Justification: See Supplement Sec. B and code is also provided separately for reproducibility.

\item {\bf Open access to data and code}
    \item[] Question: Does the paper provide open access to the data and code, with sufficient instructions to faithfully reproduce the main experimental results, as described in supplemental material?
    \item[] Answer: \answerYes{} 
    \item[] Justification: Code is provided separately for reproducibility and all datasets used are publicly available.

\item {\bf Experimental setting/details}
    \item[] Question: Does the paper specify all the training and test details (e.g., data splits, hyperparameters, how they were chosen, type of optimizer) necessary to understand the results?
    \item[] Answer: \answerYes{}{} 
    \item[] Justification: See supplement Sec. B

\item {\bf Experiment statistical significance}
    \item[] Question: Does the paper report error bars suitably and correctly defined or other appropriate information about the statistical significance of the experiments?
    \item[] Answer: \answerNo{}{} 
    \item[] Justification: The relevant numerical values are obtained by averaging over at least 10 seeds, usually 25 for stability. The exact numerical values are unimportant and only qualitative behavior is relevant to our conclusions. Further, repeating these experiments for 5 seeds for example would require training over $5\times650\times4$ models which is computationally prohibitive.

\item {\bf Experiments compute resources}
    \item[] Question: For each experiment, does the paper provide sufficient information on the computer resources (type of compute workers, memory, time of execution) needed to reproduce the experiments?
    \item[] Answer: \answerYes{} 
    \item[] Justification: See supplement Sec. B
    
\item {\bf Code of ethics}
    \item[] Question: Does the research conducted in the paper conform, in every respect, with the NeurIPS Code of Ethics \url{https://neurips.cc/public/EthicsGuidelines}?
    \item[] Answer: \answerYes{}{} 
    \item[] Justification: There are no major ethical concerns

\item {\bf Broader impacts}
    \item[] Question: Does the paper discuss both potential positive societal impacts and negative societal impacts of the work performed?
    \item[] Answer: \answerNA{} 
    \item[] Justification: The work is intended to better understand uncertainty quantification capabilities of Deep Ensembles and does not have any major societal impacts.
    
\item {\bf Safeguards}
    \item[] Question: Does the paper describe safeguards that have been put in place for responsible release of data or models that have a high risk for misuse (e.g., pre-trained language models, image generators, or scraped datasets)?
    \item[] Answer: \answerNA{}{} 
    \item[] Justification: The work is intended to better understand uncertainty quantification capabilities of Deep Ensembles and does not have any major societal impacts

\item {\bf Licenses for existing assets}
    \item[] Question: Are the creators or original owners of assets (e.g., code, data, models), used in the paper, properly credited and are the license and terms of use explicitly mentioned and properly respected?
    \item[] Answer: \answerYes{}{} 
    \item[] Justification: See Sec.~\ref{sec:app:experiments}.

\item {\bf New assets}
    \item[] Question: Are new assets introduced in the paper well documented and is the documentation provided alongside the assets?
    \item[] Answer: \answerNA{} 
    \item[] Justification: No new assets.

\item {\bf Crowdsourcing and research with human subjects}
    \item[] Question: For crowdsourcing experiments and research with human subjects, does the paper include the full text of instructions given to participants and screenshots, if applicable, as well as details about compensation (if any)? 
    \item[] Answer: \answerNA{} 
    \item[] Justification: No crowdsourcing used.

\item {\bf Institutional review board (IRB) approvals or equivalent for research with human subjects}
    \item[] Question: Does the paper describe potential risks incurred by study participants, whether such risks were disclosed to the subjects, and whether Institutional Review Board (IRB) approvals (or an equivalent approval/review based on the requirements of your country or institution) were obtained?
    \item[] Answer: \answerNA{} 
    \item[] Justification: No crowdsourcing nor research with human subjects.

\item {\bf Declaration of LLM usage}
    \item[] Question: Does the paper describe the usage of LLMs if it is an important, original, or non-standard component of the core methods in this research? Note that if the LLM is used only for writing, editing, or formatting purposes and does \emph{not} impact the core methodology, scientific rigor, or originality of the research, declaration is not required.
    \item[] Answer: \answerNA{} 
    \item[] Justification: LLMs were not used.

\end{enumerate}
\end{document}